\documentclass[twocolumn,conference]{IEEEtran}
\usepackage[T1]{fontenc}

\usepackage{xcolor}
\usepackage{amsbsy}
\usepackage{amssymb}
\usepackage{graphicx}

\ifCLASSINFOpdf

\else

\fi

\usepackage{multirow}
\usepackage{float}
\usepackage[export]{adjustbox}
\usepackage{xcolor}
\usepackage{caption}
\usepackage{subcaption}
\usepackage{afterpage}
\usepackage[labelsep=period, figurename=Figure]{caption}

\usepackage{amsmath}
\usepackage{verbatim}
\usepackage{graphicx}

\usepackage{tikz}
\usepackage{textcomp}
\usepackage{lipsum}
\usepackage{comment}
\usepackage{algpseudocode}
\usepackage{amsmath}
\usepackage{algorithm}
\usepackage{cite}
\usepackage{graphicx}
\usepackage{textcomp}
\usepackage{xcolor}
\usepackage{amsmath,amssymb,amsfonts}
\usepackage{booktabs}
\usepackage{enumitem}
\setlist[enumerate,1]{label={\arabic*.}}
\makeatother

\begin{document}

\addtolength{\textfloatsep}{-2.19pt}
\setlength{\columnsep}{0.21in}  

%

\title{Adaptive Pareto-Optimal Token Merging for Edge Transformer Models in Semantic Communication}

\author{
    \IEEEauthorblockN{Omar Erak$^{*}$, Omar Alhussein$^{*}$, Hatem Abou-Zeid$^{\dagger}$, and Mehdi Bennis$^{\ddagger}$}
        \IEEEauthorblockA{$^{*}$KU 6G Research Centre, College of Computing and Mathematical Sciences, Khalifa University, UAE\\ $^{\dagger}$Department of Electrical and Software Engineering, University of Calgary, Canada\\
                      $^{\ddagger}$Centre for Wireless Communications, University of Oulu, Finland\\
                      Emails: omarerak@ieee.org, omar.alhussein@ku.ac.ae, hatem.abouzeid@ucalgary.ca, mehdi.bennis@oulu.fi}
}

\maketitle

\begin{abstract}

Large-scale transformer models have emerged as a powerful tool for semantic communication systems, enabling edge devices to extract rich representations for robust inference across noisy wireless channels. However, their substantial computational demands remain a major barrier to practical deployment in resource-constrained 6G networks. In this paper, we present a training-free framework for adaptive token merging in pretrained vision transformers to jointly reduce inference time and transmission resource usage. We formulate the selection of per-layer merging proportions as a multi-objective optimization problem to balance accuracy and computational cost. We employ Gaussian process-based Bayesian optimization to construct a Pareto frontier of optimal configurations, enabling flexible runtime adaptation to dynamic application requirements and channel conditions. Extensive experiments demonstrate that our method consistently outperforms other baselines and achieves significant reductions in floating-point operations while maintaining competitive accuracy across a wide range of signal-to-noise ratio (SNR) conditions. Additional results highlight the effectiveness of adaptive policies that adjust merging aggressiveness in response to channel quality, providing a practical mechanism to trade off latency and semantic fidelity on demand. These findings establish a scalable and efficient approach for deploying transformer-based semantic communication in future edge intelligence systems.

\end{abstract}



\renewcommand{\thefootnote}{}
\footnotetext{An expanded version of this article can be found in\cite{erak2025adaptive}.}
\renewcommand{\thefootnote}{\arabic{footnote}}
\IEEEpeerreviewmaketitle

\section{Introduction}
The rapid evolution of wireless networks is fundamentally transforming how data and intelligence are delivered, processed, and experienced at the network edge. As we look towards sixth-generation (6G) systems, the emphasis is shifting from sheer connectivity and raw data throughput towards enabling intelligent, context-aware services that adapt in real time to users' needs and environmental dynamics \cite{6GVision}. In this new paradigm, the concept of semantic communication, extracting and transmitting informative features that are directly relevant to downstream tasks, has emerged as a key enabler for bandwidth and energy-efficient edge intelligence \cite{chaccour2024less}.

Much of the existing work on semantic communication focuses on optimizing the delivery of feature representations or compressed signals, usually through compact convolutional neural networks or task-specific autoencoders to reduce the amount of data transmitted across the network \cite{bourtsoulatze2019deep, erak2024contrastive}. However, recent breakthroughs in deep learning, particularly with the rise of transformer architectures \cite{vaswani2017attention}, have demonstrated remarkable advances in extracting semantically rich representations from high-dimensional multi-modal data such as images, audio, and text. Despite their promise, transformers remain computationally and communicatively intensive, which makes their practical deployment at the wireless edge challenging due to strict constraints on computation, memory, and uplink bandwidth \cite{letaief2021edge}.

While there has been increasing interest in developing efficient deep neural networks for edge intelligence and 6G applications \cite{10279462}, the question of how to enable resource adaptive, transformer-based semantic communication for edge-to-cloud scenarios remains largely underexplored. In particular, there is a critical need for frameworks that can dynamically balance task accuracy with inference and communication efficiency, without the overhead of retraining or extensive manual tuning each time network or task conditions change.

In this paper, we propose a training-free, multi-objective optimization framework for transformer-based semantic communication at the wireless edge.  Our approach enables an edge device to process raw data using a pretrained transformer model, dynamically merging and reducing semantic tokens as they propagate through the model’s layers. The resulting compact set of tokens is then transmitted over a joint source-channel coded (JSCC) wireless link to a server for downstream task execution. We formulate the token merging process as a multi-objective optimization problem, jointly maximizing task performance and minimizing inference time. By employing Bayesian optimization \cite{frazier2018tutorialbayesianoptimization} and constructing a Pareto front of optimal merging configurations, our method empowers users or network controllers to flexibly select the most suitable operating point for any given scenario, trading off accuracy and efficiency in real time, and doing so entirely without retraining.

To the best of our knowledge, this is the first work to address efficient, adaptive transformer-based semantic communication in a training-free fashion for 6G edge intelligence. While the Bayesian optimization process requires offline evaluations to build the Pareto front, no retraining or weight updates of the transformer model are performed. Our experiments demonstrate that the proposed approach achieves significant reductions in computational cost and inference time at the edge, while maintaining competitive task accuracy. The framework also offers practical, plug-and-play adaptability to changing network or application requirements.

The remainder of this paper is organized as follows: Section II reviews related work on transformer-based semantic communication, and token reduction techniques. Section III introduces our system model and  Section IV presents the problem statement. Section V details the proposed token merging and multi-objective optimization framework. Section VI describes the experimental setup and discusses the results. Section VII concludes our work and outlines future directions.

\section{Related Work}

Qiao et al. introduced Token Communications (TokCom) \cite{qiao2025tokencommunicationsunifiedframework}, which defines semantic-level units based on transformer tokens, and uses pretrained models to generate and transmit context-aware tokens. Their approach shows improvements compared to previous work, however, improving the efficiency of the transformer model is not explored. Similarly, Xie et al. introduced large model-empowered semantic communication systems based on pretrained transformers \cite{10599117}, their results show improvements compared to previous baselines, however they highlight the need for further research to improve the deployment and usage of these computationally demanding models. Devoto et al. proposed an adaptive, trainable token pruning mechanism within a transformer and JSCC pipeline \cite{devoto2024adaptivesemantictokenselection}. Their work requires retraining and is focused on dropping tokens entirely instead of merging them. 

Bolya et al. introduced Token Merging (ToMe) \cite{bolya2023token}, a training-free mechanism that merges similar transformer tokens during inference using a matching algorithm. Unlike pruning, ToMe reduces token redundancy by combining rather than dropping, achieving up to 2× inference speedup with minimal accuracy loss on Vision Transformers (ViTs). ToMe merges tokens based on similarity but uses a fixed merging schedule per layer and utilizes averaging to combine the tokens. Liu et al. proposed Adaptive Sparse ViT (AS-ViT) \cite{10.24963/ijcai.2023/136}, that introduces learnable thresholds within self-attention layers to prune unimportant tokens dynamically during inference. Their scheme, trained with a budget-aware loss, can adapt pruning per sample and achieve significant speedups with minor accuracy degradation. However, this method requires retraining and focuses on token pruning instead of merging. Bayesian optimization for multi-objective latency-accuracy trade-offs has also been investigated in the context of neural architecture search \cite{eriksson2021latencyaware}, but this line of work has not been explored for the dynamic selection or merging of transformer tokens.

We present a training-free, multi-objective token merging framework explicitly designed for resource-constrained edge devices. Like ToMe, our framework applies merging during inference without any retraining. However, unlike prior methods, we introduce norm-based weighted merging to improve representation quality over simple averaging; while avoiding fixed drop schedules or pruning mechanisms. Furthermore, we frame token merging as a multi-objective optimization problem, constructing a Pareto frontier of configurations that trade off compute, latency, and task accuracy. This allows runtime, user-driven selection of optimal behaviors without retraining, delivering flexible, plug-and-play adaptability for transformer-based semantic communication.
\begin{figure*}[t]
    \centering
    \includegraphics[width=0.95\textwidth]{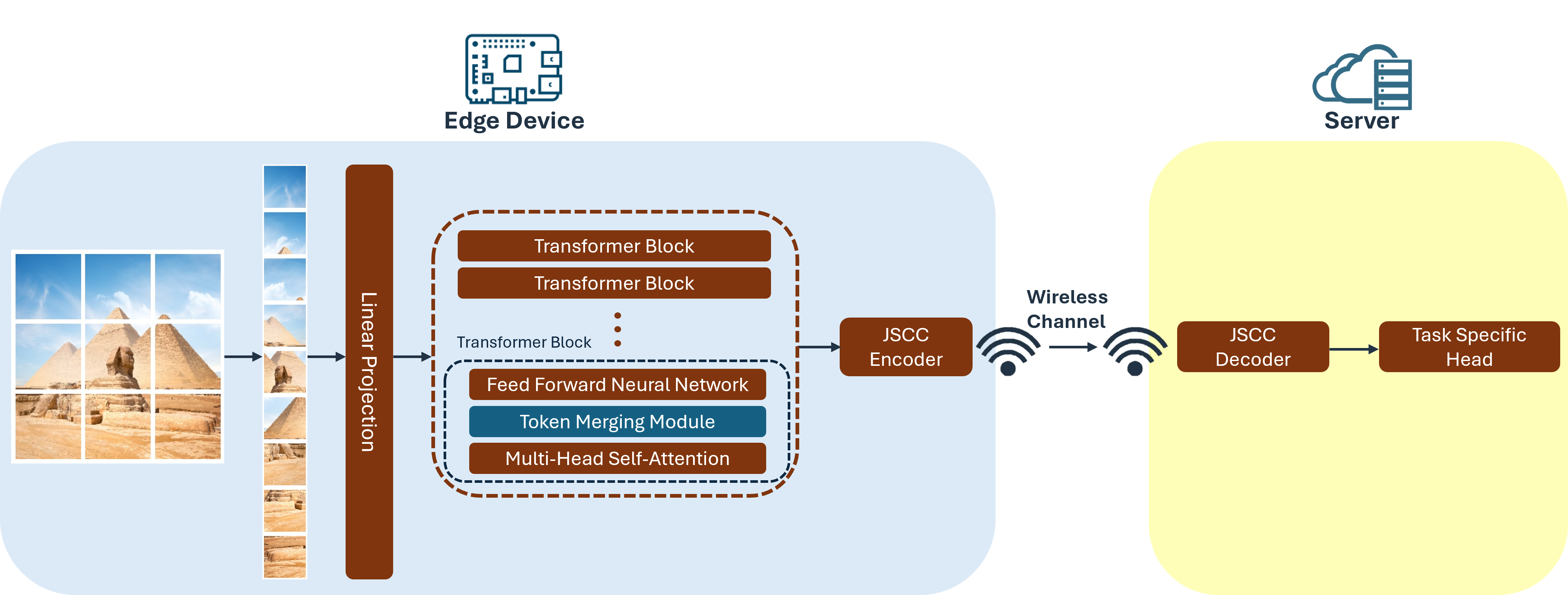}
    \caption{Overview of the proposed edge-to-cloud semantic communication system. An input image is partitioned into patches and projected into embeddings, which are processed by a pretrained transformer encoder augmented with a training-free token merging module. The resulting semantic tokens are compressed by a JSCC encoder and transmitted over a noisy wireless channel. The server reconstructs the tokens using a JSCC decoder and performs task-specific inference.}
    \label{fig:system_model}
\end{figure*}

\section{System Model}

We consider an edge-to-cloud semantic communication system where an edge device transmits compressed semantic representations of sensory data to a remote server over a noisy wireless channel. The server subsequently performs a downstream inference task using the received semantic tokens. The system model is summarized in Fig.~\ref{fig:system_model}.

Let $\boldsymbol{x} \in \mathbb{R}^{H \times W \times C}$ denote the input data (e.g., an image of height $H$, width $W$, and $C$ channels) observed at the edge device. The edge applies a patch embedding operation, denoted $\mathcal{E}$, which partitions $\boldsymbol{x}$ into $N$ non-overlapping patches and projects each patch into a $d$-dimensional embedding, resulting in an initial set of semantic tokens as follows
\begin{equation}
    \boldsymbol{Z}_0 = \mathcal{E}(\boldsymbol{x}) = [\boldsymbol{z}_1^{(0)}, \ldots, \boldsymbol{z}_N^{(0)}], \qquad \boldsymbol{z}_i^{(0)} \in \mathbb{R}^d.
\end{equation}
These tokens are processed by a pretrained transformer encoder $\mathcal{T}_\theta$ with fixed parameters $\theta$.

The tokens are then propagated through $L$ transformer layers. At each layer $\ell = 1, \ldots, L$, a training-free token merging operation, denoted $\mathcal{K}_{p_\ell}(\cdot)$, is applied. Here, $p_\ell$ represents the percentage of tokens to be merged at layer $\ell$, reducing the number of tokens to $N_\ell \leq N_{\ell-1}$, given by
\begin{equation}
    \boldsymbol{Z}_\ell = \mathcal{K}_{p_\ell}\left(\mathcal{T}^{(\ell)}_\theta(\boldsymbol{Z}_{\ell-1})\right), \qquad \boldsymbol{Z}_\ell \in \mathbb{R}^{N_\ell \times d},
\end{equation}
where $N_0 = N$.

After the final transformer layer, the merged token set $\boldsymbol{Z}_L \in \mathbb{R}^{N_L \times d}$ is mapped to a channel symbol vector using a JSCC encoder $\mathcal{J}$ as follows
\begin{equation}
    \boldsymbol{s} = \mathcal{J}(\boldsymbol{Z}_L), \qquad \boldsymbol{s} \in \mathbb{C}^{q},
\end{equation}
where $q$ is the number of channel uses.

The channel input $\boldsymbol{s}$ is normalized such that its average power per channel use satisfies $\mathbb{E}[\|\boldsymbol{s}\|^2]/q = 1$. The encoded signal is transmitted over an additive white Gaussian noise (AWGN) channel, given by
\begin{equation}
    \boldsymbol{s}' = \boldsymbol{s} + \boldsymbol{n},
\end{equation}
where $\boldsymbol{n} \sim \mathcal{CN}(0, \sigma^2 \boldsymbol{I}_q)$ is the AWGN vector with variance $\sigma^2$. The signal-to-noise ratio (SNR) at the receiver, expressed in decibels (dB), is given by
\begin{equation}
    \mathrm{SNR_{dB}} = 10 \log_{10} \left( \frac{1}{\sigma^2} \right).
\end{equation}

At the server, a JSCC decoder $\mathcal{D}$ reconstructs the semantic tokens as follows
\begin{equation}
    \hat{\boldsymbol{Z}}_L = \mathcal{D}(\boldsymbol{s}').
\end{equation}

The downstream inference module $f_\phi$, with parameters $\phi$, uses the recovered tokens to perform the task of interest, e.g., classification, given by
\begin{equation}
    \hat{y} = f_\phi(\hat{\boldsymbol{Z}}_L),
\end{equation}
where $\hat{y}$ is the predicted label or output.

\section{Problem Statement}

Transformer-based models have set the state-of-the-art across a wide range of tasks, but their high computational complexity presents a major obstacle for edge deployment. In particular, the self-attention mechanism in each transformer layer operates with a complexity of $\mathcal{O}(N^2 d)$ \cite{keles2023computational}. As a result, inference with large $N$ is prohibitively expensive for resource-constrained edge devices. Therefore, if $N$ can be reduced during inference, significant savings in computation can be achieved without retraining.

 Let $\boldsymbol{p} = [p_1, \ldots, p_L]$ denote the vector of merging parameters. For a given $\boldsymbol{p}$, let $A(\boldsymbol{p})$ denote the downstream task accuracy, and let $F(\boldsymbol{p})$ denote the total number of floating point operations (FLOPs) required for inference. 
 
 The mappings $A(\boldsymbol{p})$ and $F(\boldsymbol{p})$ are non-convex and do not admit closed-form expressions, as they depend on complex interactions within the model and input data distribution. Furthermore, the search space of possible merging policies $\mathcal{P}$ is high-dimensional, and evaluating each candidate configuration is computationally expensive due to the need for a full model forward pass and accuracy assessment.  As a result, this problem is well-suited for black-box, sample-efficient optimization methods.

 We thus formulate the following multi-objective optimization problem:
\begin{equation}
    \max_{\boldsymbol{p} \in \mathcal{P}} \big[ A(\boldsymbol{p}),\; -F(\boldsymbol{p}) \big],
\end{equation}
where the goal is to characterize the set of Pareto-optimal tradeoffs-configurations for which no objective can be improved without degrading the other. Each point on the Pareto front represents an optimal tradeoff between accuracy and computation, allowing a decision-maker to select a configuration that best meets the real-time requirements of the application. This formulation enables runtime adaptation in a training-free manner, leveraging only the flexibility of the token merging mechanism during inference.

\section{Methodology}

In this section, we first describe our training-free token merging mechanism for reducing transformer inference complexity at the edge. We then present our multi-objective Bayesian optimization framework for efficiently identifying Pareto-optimal merging configurations that balance accuracy and computational cost.

\subsection{Training-Free Token Merging Mechanism}
At each transformer layer $\ell$, we apply a training-free token merging operation to adaptively reduce the number of tokens, thereby lowering computational complexity and inference latency. Recall that $\boldsymbol{Z}_{\ell-1} = [\boldsymbol{z}_1^{(\ell-1)}, \ldots, \boldsymbol{z}_{N_{\ell-1}}^{(\ell-1)}] \in \mathbb{R}^{N_{\ell-1}\times d}$ denotes the input tokens to layer $\ell$. To determine which tokens to merge, we first compute semantic similarity using the Value vectors from self-attention. The Value matrix at layer $\ell$ is given by
\begin{equation}
\boldsymbol{V}^{(\ell)} = \boldsymbol{Z}_{\ell-1}\,\boldsymbol{W}_V^{(\ell)},
\end{equation}
where $\boldsymbol{W}_V^{(\ell)}\in\mathbb{R}^{d\times d}$ is the value projection matrix. We define the index sets of alternating tokens:
\begin{equation}
\mathcal{A}_\ell = \{1,3,5,\ldots\},\quad \mathcal{B}_\ell = \{2,4,6,\ldots\}.
\end{equation}
We refer to tokens in $\mathcal{A}_\ell$ as \emph{sources} and those in $\mathcal{B}_\ell$ as \emph{destinations} for the merging operation.

For each index $a\in\mathcal{A}_\ell$, we compute the cosine similarity with every $b\in\mathcal{B}_\ell$:
\begin{equation}
s_{ab}^{(\ell)} = \frac{\langle \boldsymbol{v}_a^{(\ell)},\,\boldsymbol{v}_b^{(\ell)}\rangle}{\|\boldsymbol{v}_a^{(\ell)}\|_2\,\|\boldsymbol{v}_b^{(\ell)}\|_2}.
\end{equation}
We select the $r_\ell = \lfloor p_\ell N_{\ell-1}\rfloor$ highest-scoring assignments of sources to destinations, where $p_\ell\in[0,0.3]$ is the merge proportion. Note that multiple tokens in $\mathcal{A}_\ell$ may be assigned to the same destination in $\mathcal{B}_\ell$. Let $\mathcal{M}_\ell\subset\mathcal{B}_\ell$ denote the set of indices of destination tokens that receive at least one merge assignment. For each index $m\in\mathcal{M}_\ell$, let $\mathcal{S}_m\subset\mathcal{A}_\ell$ denote the set of indices of source tokens assigned to merge into token $m$. The merged embedding corresponding to destination index $m$ is computed as
\begin{equation}
\boldsymbol{u}_m^{(\ell)}=
\frac{
\|\boldsymbol{z}_m^{(\ell-1)}\|_2\,\boldsymbol{z}_m^{(\ell-1)}
+
\sum_{s\in\mathcal{S}_m}\|\boldsymbol{z}_s^{(\ell-1)}\|_2\,\boldsymbol{z}_s^{(\ell-1)}
}{
\|\boldsymbol{z}_m^{(\ell-1)}\|_2
+
\sum_{s\in\mathcal{S}_m}\|\boldsymbol{z}_s^{(\ell-1)}\|_2
+\varepsilon
},
\end{equation}
where $\varepsilon$ is a small constant added for numerical stability. Finally, let $\mathcal{R}_\ell$ denote the indices of tokens not involved in merging (i.e., retained without modification). The output tokens after merging are given by
\begin{equation}
\boldsymbol{Z}_\ell=
\bigl[
\{\boldsymbol{z}_r^{(\ell-1)}\}_{r\in\mathcal{R}_\ell},
\{\boldsymbol{u}_m^{(\ell)}\}_{m\in\mathcal{M}_\ell}
\bigr],
\end{equation}
where $N_\ell=|\mathcal{R}_\ell|+|\mathcal{M}_\ell|$ denotes the total number of tokens after merging. The overall procedure applied across all transformer layers is summarized in Algorithm~\ref{alg:token_merging_all_layers}.

\begin{algorithm}[]
\footnotesize
\caption{Token Merging Across Transformer Layers}
\label{alg:token_merging_all_layers}
\begin{algorithmic}[1]
\Require Initial embeddings $\boldsymbol{Z}_0$, merge proportions $p_\ell$
\Ensure Final embeddings $\boldsymbol{Z}_L$
\For{$\ell=1,\ldots,L$}
    \State Compute Value matrix $\boldsymbol{V}^{(\ell)}$
    \State Split indices: $\mathcal{A}_\ell$, $\mathcal{B}_\ell$
    \For{each $a\in\mathcal{A}_\ell$}
        \State Compute $s_{ab}$ for all $b$
        \State Record best $b$ and score
    \EndFor
    \State Collect and sort all scores
    \State Select top $r_\ell$ pairs based on $p_\ell$
    \State Get $\mathcal{M}_\ell$ and sets $\mathcal{S}_m$
    \For{each $m\in\mathcal{M}_\ell$}
        \State Norm-weighted merge
    \EndFor
    \State Concatenate retained and merged tokens
\EndFor
\end{algorithmic}
\end{algorithm}

\subsection{Multi-Objective Bayesian Optimization for Pareto-Optimal Merging}

Let $\boldsymbol{p} = [p_1, \ldots, p_L]$ denote the vector of merging proportions across $L$ transformer layers, with each $p_i \in [0,0.3]$. For a given configuration $\boldsymbol{p}$, let $A(\boldsymbol{p})$ and $F(\boldsymbol{p})$ represent the downstream task accuracy and the total number of floating-point operations required for inference, respectively. The objective is to identify configurations that optimally trade off these conflicting goals. Formally, the Pareto front is defined as follows
\begin{equation}
\mathcal{F}^* = \bigl\{\boldsymbol{p}\in\mathcal{P}\,\big|\,
\nexists\,\boldsymbol{p}'\in\mathcal{P}:
A(\boldsymbol{p}') \ge A(\boldsymbol{p}) \wedge F(\boldsymbol{p}') \le F(\boldsymbol{p})
\bigr\},
\end{equation}
where $\mathcal{P}$ denotes the feasible domain of merging configurations. In this definition, at least one of the inequalities must be strict to ensure that configurations identical in both objectives do not dominate each other.

Because the mappings $A(\boldsymbol{p})$ and $F(\boldsymbol{p})$ are a black-box, non-convex, high-dimensional, and costly to evaluate, we adopt Gaussian process (GP) Bayesian optimization to efficiently guide the search for Pareto-optimal configurations \cite{rasmussen2003gaussian, frazier2018tutorialbayesianoptimization}. In this approach, each objective is modeled as a realization of a GP prior:
\begin{equation}
    f(\boldsymbol{p}) \sim \mathcal{GP}\bigl(\mu(\boldsymbol{p}),\,k(\boldsymbol{p},\boldsymbol{p}')\bigr),
\end{equation}
where $\mu(\boldsymbol{p})$ is the prior mean function and $k(\boldsymbol{p},\boldsymbol{p}')$ is the covariance kernel. We choose the mean function to be constant and employ a Matérn-5/2 kernel with Automatic Relevance Determination (ARD) \cite{genton2001classes}, defined in its simplified form as
\begin{equation}
    k(\boldsymbol{p},\boldsymbol{p}') = \sigma_f^2 \Bigl(1 + \sqrt{5}\,r + \frac{5}{3}\,r^2\Bigr)\,\exp\bigl(-\sqrt{5}\,r \bigr),
\end{equation}
with
\[
    r = \sqrt{\sum_{i=1}^{L}\frac{\bigl(p_i - p_i'\bigr)^2}{\ell_i^2}}.
\]
Here, $\sigma_f^2$ denotes the signal variance, which governs the overall variability of the function, while the length-scale parameters $\{\ell_i\}$ reflect the sensitivity of each merging proportion dimension to variations in the objective. Smaller length-scales indicate that small changes in $p_i$ can lead to rapid changes in the output, whereas larger length-scales imply smoother, less sensitive dependence. The ARD formulation thus enables the model to automatically learn which layers have the greatest impact on performance and computational cost. All kernel hyperparameters, including $\sigma_f^2$ and $\{\ell_i\}$, are estimated by maximizing the marginal likelihood of the observed data. To improve robustness and mitigate overfitting, Gamma priors are applied to the kernel scale and the observation noise variance, and additional priors are placed over the inverse squared length-scales \cite{akiba2019optunanextgenerationhyperparameteroptimization}.

Candidate configurations are selected by maximizing an acquisition function computed from the GP posterior predictive distributions. In the multi-objective setting, we adopt the logarithm of the Expected Hypervolume Improvement (EHVI) criterion~\cite{daulton2020differentiable}, which quantifies the expected gain in the hypervolume dominated by the Pareto front when evaluating a candidate configuration. This criterion balances exploration of uncertain regions with exploitation of promising areas, enabling efficient approximation of the Pareto-optimal trade-offs. After a fixed budget of evaluations, the set of non-dominated configurations among those actually sampled provides an empirical approximation of the Pareto front as follows
\begin{equation}
\widehat{\mathcal{F}} = \bigl\{\boldsymbol{p}\in \mathcal{S}_T\,\big|\,
\nexists\,\boldsymbol{p}'\in\ \mathcal{S}_T:
A(\boldsymbol{p}') \ge A(\boldsymbol{p}) \wedge F(\boldsymbol{p}') \le F(\boldsymbol{p})
\bigr\},
\end{equation}
where at least one of the inequalities must be strict, and $\mathcal{S}_T$ denotes the set of configurations evaluated up to iteration $T$. After constructing the empirical Pareto front, the edge device can select merging configurations from the Pareto front that either maximize accuracy under a latency constraint or dynamically balance performance and computational cost in response to changing conditions.

\section{Experimental Setup and Evaluation}

\subsection{Experimental Setup}
\label{sec:experimental_setup}

We evaluate our framework on the ImageNet-1k validation set using a ViT-Base/16 model pretrained with masked autoencoding \cite{5206848, dosovitskiy2020image}. All transformer weights are frozen during inference. For each transformer layer, a merging proportion $p_\ell \in [0,0.3]$ specifies the fraction of tokens to merge. Gaussian process Bayesian optimization with a Matérn-5/2 kernel and ARD is conducted for 500 trials to identify Pareto-optimal configurations. The optimization is performed on a randomly sampled subset of 1000 validation images to reduce computational cost, while final evaluations are conducted on the full validation set. Each trial records top-1 classification accuracy and total FLOPs. The resulting configurations are then evaluated under both noiseless inference and noisy channel conditions. Specifically, an AWGN channel is simulated across SNR levels from -10 to 25~dB. We employ a lightweight convolutional JSCC encoder-decoder architecture similar to \cite{bourtsoulatze2019deep}, trained separately on ImageNet embeddings. Baselines include the uncompressed model with no merging, uniform merging corresponding to the ToMe procedure applied with fixed proportions between 10\% and 30\%, and random merging schedules sampled uniformly. We report accuracy versus FLOPs trade-offs in the noiseless setting, end-to-end accuracy as a function of SNR, and throughput measured in images per second on an NVIDIA RTX 4080 GPU. Visual examples are also provided to illustrate the qualitative impact of token merging. \footnote{The source code, models and results will be made available at https://github.com/OmarErak/adaptive-token-merging-semcom}

\begin{figure}
    \centering
    \includegraphics[width=0.9\linewidth]{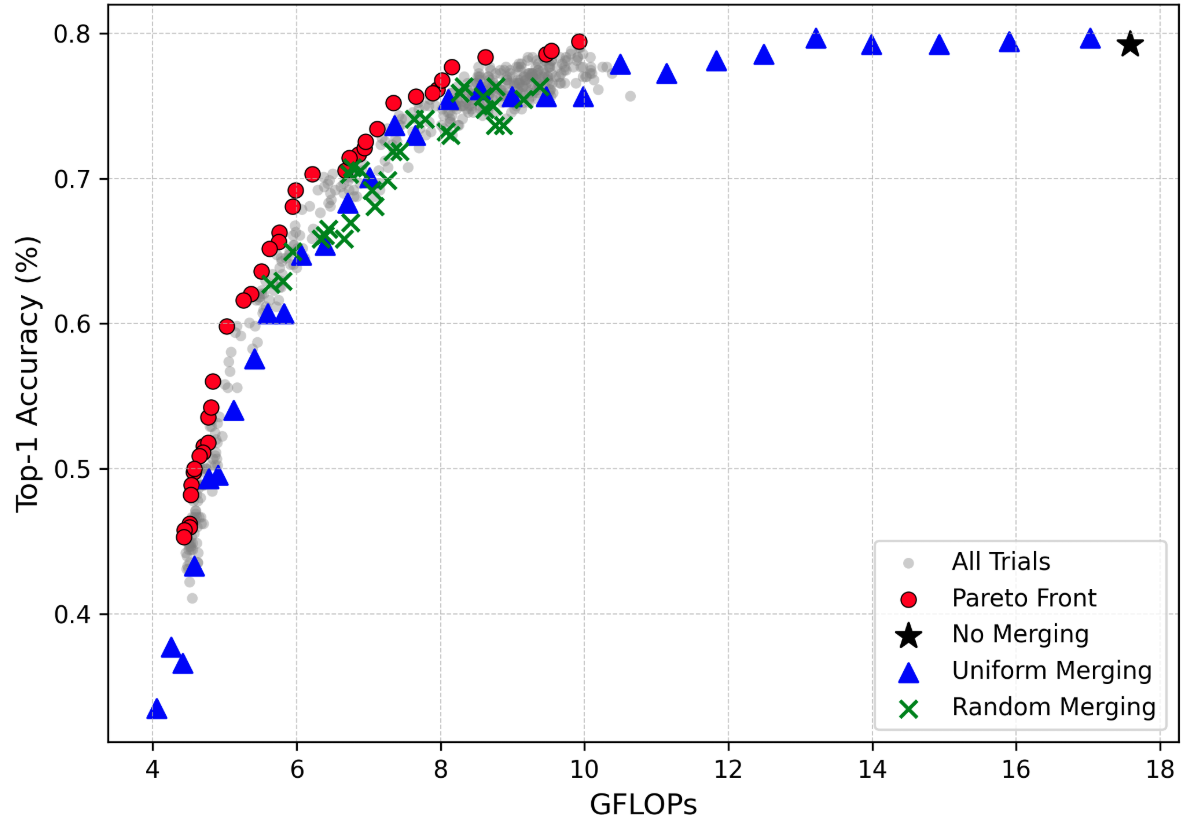}
    \caption{Accuracy vs. GFLOPs trade-off across all evaluated configurations. Gray dots indicate all sampled configurations during Bayesian optimization. Red circles denote Pareto-optimal configurations discovered by our method.}
    \label{fig:exp1}
\end{figure}

\begin{table}[]
\vspace*{0.1in}  
\setlength{\tabcolsep}{4pt}
\centering
\caption{Performance of Selected Configurations Under SNR = 20 dB, Demonstrating Application-Specific Adaptability}
\begin{tabular}{@{}lccc@{}}
\toprule
\textbf{Scenario} & \textbf{Accuracy (\%)} & \textbf{GFLOPs} & \textbf{Throughput (img/s)} \\
\midrule
S1 & 79.0 & 9.9 & 423.8 \\
S2 & 75.0 & 7.3 & 521.9 \\
S3 & 68.5 & 6.0 & 551.3 \\
\bottomrule
\end{tabular}
\label{tab:experiment3a_results}
\end{table}

\subsection{Experimental Evaluation}

\begin{figure}[]
    \centering
    \includegraphics[width=0.8\linewidth]{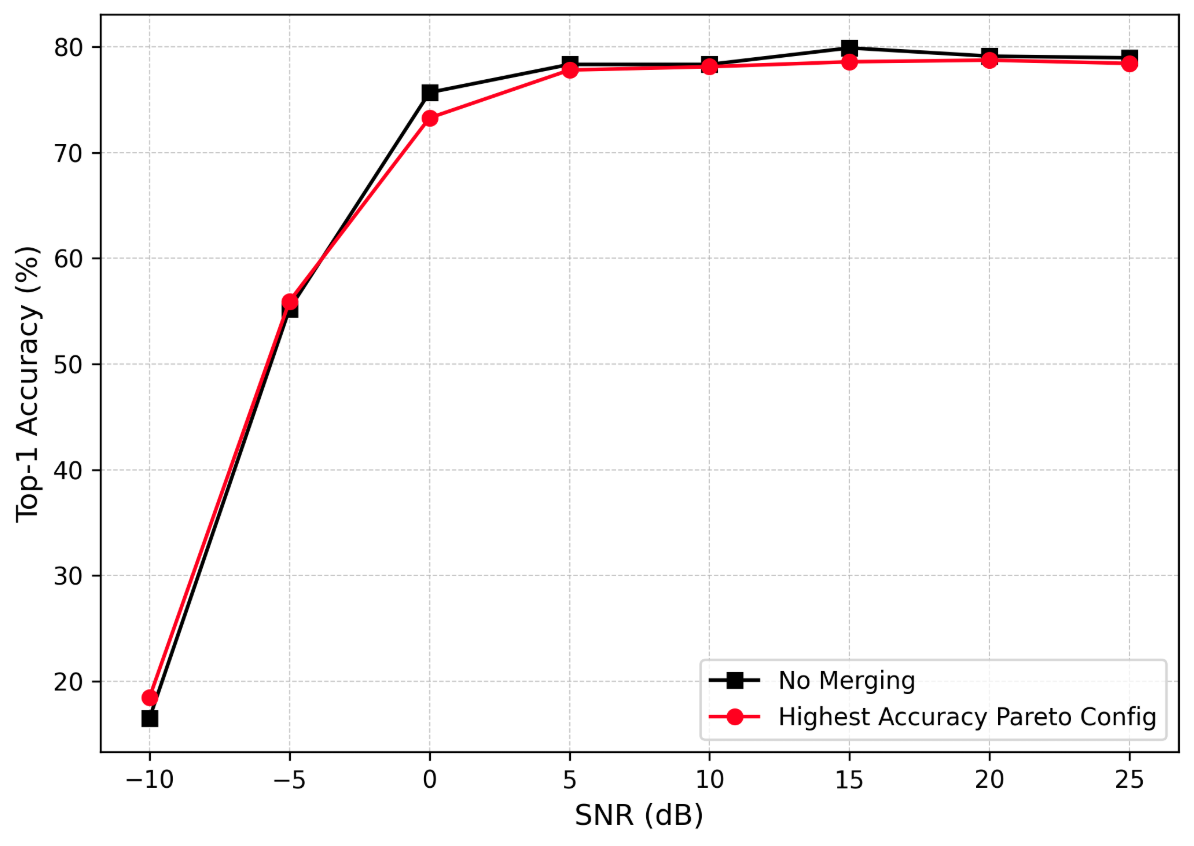}
    \caption{Top-1 accuracy across varying SNR levels for the highest accuracy Pareto configuration compared to the uncompressed model.}
    \label{fig:exp2}
\end{figure}
\begin{figure}[]
    \centering
    \includegraphics[width=0.8\linewidth]{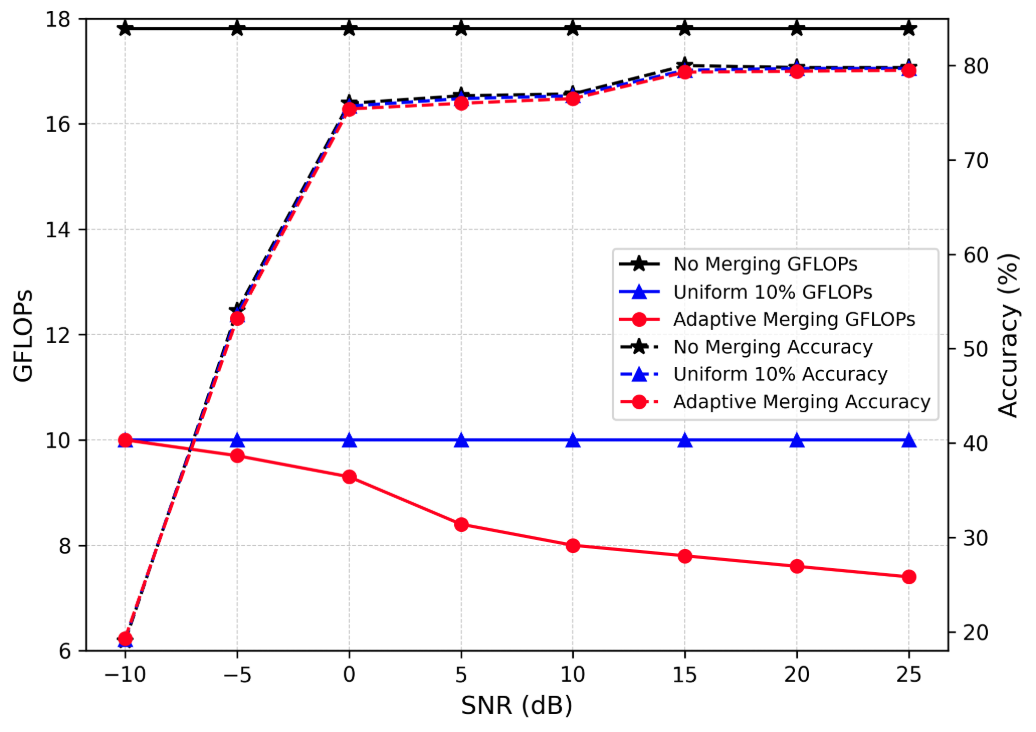}
    \caption{GFLOPs and accuracy versus SNR for the different merging strategies}
    \label{fig:exp3}
\end{figure}

Fig. \ref{fig:exp1} illustrates the trade-off between computational complexity and task accuracy achieved by the proposed multi-objective Bayesian optimization framework compared to baseline strategies. The Pareto front consistently dominates uniform and random merging across the accuracy-FLOP spectrum, highlighting the advantage of optimizing per-layer merging proportions rather than applying a fixed or random compression schedule.

Table \ref{tab:experiment3a_results} summarizes the performance of selected configurations under an SNR of 20 dB. Scenario 1 corresponds to the configuration that maximizes accuracy. Scenario 2 demonstrates a configuration that minimizes FLOPs subject to maintaining accuracy above 75\%. Scenario 3 reflects an operating point prioritizing throughput under a constraint of at most 6 GFLOPs. The results show that our proposed framework enables flexible, application-specific selection of operating points, achieving substantial reductions in GFLOPs while preserving competitive accuracy and adhering to any imposed constraints.

\begin{figure}[t]
    \centering
    \includegraphics[width=0.85\linewidth]{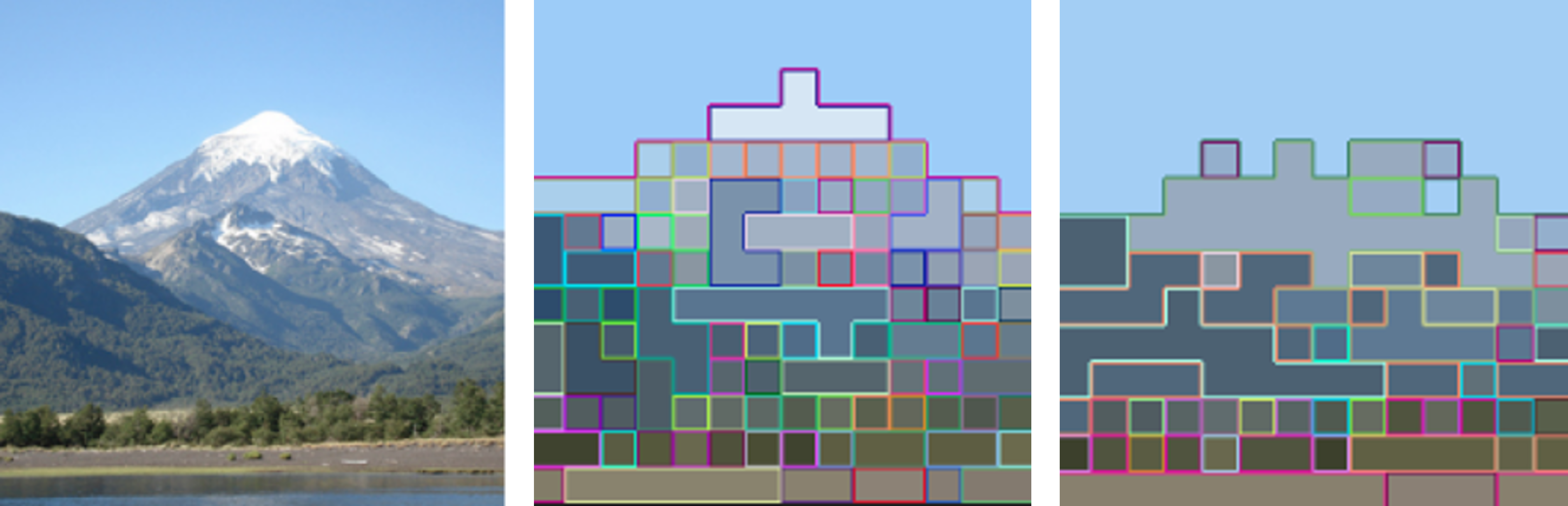}
    \caption{
    Left: Original image. Center: High-accuracy configuration sampled from the Pareto front. Right: Lower-complexity configuration from the Pareto front showing more aggressive merging.
    }
    \label{fig:token_merging_visualization}
\end{figure}

Fig. \ref{fig:exp2} further evaluates robustness to wireless channel conditions. The highest-accuracy Pareto configuration and the no merging baseline are compared under varying SNR levels. Both strategies show graceful degradation as SNR decreases, highlighting the effectiveness of adaptive token merging for efficient edge-based semantic communication. Fig. \ref{fig:exp3} highlights the adaptive policy’s ability to dynamically adjust merging aggressiveness in response to SNR. As channel conditions improve, the adaptive strategy progressively increases token merging which results in a reduction in GFLOPs while maintaining high accuracy. Unlike fixed baselines, this demonstrates the capability to flexibly scale computational cost in response to environmental variability. This property is critical for edge intelligence systems operating under dynamic resource constraints. Fig. ~\ref{fig:token_merging_visualization} shows an example visualization of token merging behavior on an ImageNet validation image. The high-accuracy configuration retains finer-grained token partitions, while the more aggressive merging configuration reduces the number of tokens and coarsens the representation. Both configurations were selected from the Pareto front, demonstrating how the framework provides interpretable and controllable trade-offs between efficiency and semantic fidelity.

\section{Conclusions and Future Directions}
This paper presented a training-free framework for adaptive token merging in transformer-based semantic communication systems targeting edge-to-cloud scenarios. By formulating token merging as a multi-objective optimization problem and leveraging Gaussian process-based Bayesian optimization, we identified Pareto-optimal configurations that balance accuracy, computational complexity, and communication efficiency. Extensive experiments demonstrated that the proposed approach consistently outperforms uniform and random merging baselines, achieving substantial reductions in FLOPs while maintaining high semantic fidelity, even under varying channel conditions. Furthermore, we showed that adaptive policies can dynamically select merging configurations in response to channel SNR, enabling flexible trade-offs between throughput and performance in real time. Future work will explore extending this framework to multimodal transformers, integrating hardware-aware objectives such as energy consumption, and applying the method to other downstream tasks including object detection, semantic segmentation and visual question answering.

\bibliographystyle{IEEEtran}

\bibliography{references}

@article{chaccour2024less,
  title={Less data, more knowledge: Building next generation semantic communication networks},
  author={Chaccour, Christina and Saad, Walid and Debbah, Merouane and Han, Zhu and Poor, H Vincent},
  journal={IEEE Commun. Surveys Tuts.},
  year={2024},
  publisher={IEEE}
}

@ARTICLE{6GVision,
  author={Saad, Walid and Bennis, Mehdi and Chen, Mingzhe},
  journal={IEEE Netw.}, 
  title={A Vision of {6G} Wireless Systems: Applications, Trends, Technologies, and Open Research Problems}, 
  year={2020},
  volume={34},
  number={3},
  pages={134-142},
  doi={10.1109/MNET.001.1900287}}

@INPROCEEDINGS{10279462,
  author={Erak, Omar and Abou-Zeid, Hatem},
  booktitle={ Proc. IEEE Int. Conf.
Commun. (ICC)},
  title={Accelerating and Compressing Deep Neural Networks for Massive MIMO CSI Feedback}, 
  year={2023},
  volume={},
  number={},
  pages={1029-1035},
  doi={10.1109/ICC45041.2023.10279462}}

@article{bourtsoulatze2019deep,
  title={Deep joint source-channel coding for wireless image transmission},
  author={Bourtsoulatze, Eirina and Kurka, David Burth and G{\"u}nd{\"u}z, Deniz},
  journal={IEEE Trans. Cogn. Commun. Netw},
  volume={5},
  number={3},
  pages={567--579},
  year={2019},
  publisher={IEEE}
}

@article{erak2024contrastive,
  title={Contrastive Learning and Adversarial Disentanglement for Task-Oriented Semantic Communications},
  author={Erak, Omar and Alhussein, Omar and Tong, Wen},
  journal={arXiv preprint arXiv:2410.22784},
  year={2024}
}

@article{letaief2021edge,
  title={Edge artificial intelligence for 6G: Vision, enabling technologies, and applications},
  author={Letaief, Khaled B and Shi, Yuanming and Lu, Jianmin and Lu, Jianhua},
  journal={IEEE J. Sel. Areas Commun},
  volume={40},
  number={1},
  pages={5--36},
  year={2021},
  publisher={IEEE}
}

@article{vaswani2017attention,
  title={Attention is all you need},
  author={Vaswani, Ashish and Shazeer, Noam and Parmar, Niki and Uszkoreit, Jakob and Jones, Llion and Gomez, Aidan N and Kaiser, {\L}ukasz and Polosukhin, Illia},
  journal={Proc. Adv. Neural Inf. Process.
Syst},
  volume={30},
  year={2017}
}

@ARTICLE{10599117,
  author={Xie, Huiqiang and Qin, Zhijin and Tao, Xiaoming and Han, Zhu},
  journal={IEEE Commun. Mag.}, 
  title={Toward Intelligent Communications: Large Model Empowered Semantic Communications}, 
  year={2025},
  volume={63},
  number={1},
  pages={69-75},
  doi={10.1109/MCOM.001.2300807}}

@article{qiao2025tokencommunicationsunifiedframework,
      title={Token Communications: A Unified Framework for Cross-modal Context-aware Semantic Communications}, 
      author={Li Qiao and Mahdi Boloursaz Mashhadi and Zhen Gao and Rahim Tafazolli and Mehdi Bennis and Dusit Niyato},
      journal={arXiv preprint arXiv:2502.12096},
      year={2025}, 
}

@article{devoto2024adaptivesemantictokenselection,
      title={Adaptive Semantic Token Selection for AI-native Goal-oriented Communications}, 
      author={Alessio Devoto and Simone Petruzzi and Jary Pomponi and Paolo Di Lorenzo and Simone Scardapane},
journal={arXiv preprint arXiv:2405.02330},
      year={2024},
}

@inproceedings{
bolya2023token,
title={Token Merging: Your ViT But Faster},
author={Daniel Bolya and Cheng-Yang Fu and Xiaoliang Dai and Peizhao Zhang and Christoph Feichtenhofer and Judy Hoffman},
booktitle={Proc. Int. Conf. Learn. Represent. },
year={2023},

}

@inproceedings{10.24963/ijcai.2023/136,
author = {Liu, Xiangcheng and Wu, Tianyi and Guo, Guodong},
title = {Adaptive sparse ViT: towards learnable adaptive token pruning by fully exploiting self-attention},
year = {2023},
isbn = {978-1-956792-03-4},
doi = {10.24963/ijcai.2023/136},
articleno = {136},
numpages = {9},
location = {Macao, P.R.China},
series = {IJCAI '23}
}

@inproceedings{
eriksson2021latencyaware,
title={Latency-Aware Neural Architecture Search with Multi-Objective Bayesian Optimization},
author={David Eriksson and Pierce I-Jen Chuang and Samuel Daulton and Peng Xia and Akshat Shrivastava and Arun Babu and Shicong Zhao and Ahmed A Aly and Ganesh Venkatesh and Maximilian Balandat},
booktitle={8th ICML Workshop on Automated Machine Learning (AutoML) },
year={2021},
}

@article{frazier2018tutorialbayesianoptimization,
      title={A Tutorial on Bayesian Optimization}, 
      author={Peter I. Frazier},
journal={arXiv preprint arXiv:1807.02811},
      year={2018},
}

@inproceedings{keles2023computational,
  title={On the computational complexity of self-attention},
  author={Keles, Feyza Duman and Wijewardena, Pruthuvi Mahesakya and Hegde, Chinmay},
  booktitle={Int. Conf. Algorithmic Learn. Theory},
  pages={597--619},
  year={2023},
  organization={PMLR}
}

@incollection{rasmussen2003gaussian,
  title={Gaussian processes in machine learning},
  author={Rasmussen, Carl Edward},
  booktitle={Summer school on machine learning},
  pages={63--71},
  year={2003},
  publisher={Springer}
}

@article{genton2001classes,
  title={Classes of kernels for machine learning: a statistics perspective},
  author={Genton, Marc G},
  journal={J. Mach. Learn. Res.},
  volume={2},
  number={Dec},
  pages={299--312},
  year={2001}
}

@misc{akiba2019optunanextgenerationhyperparameteroptimization,
      title={Optuna: A Next-generation Hyperparameter Optimization Framework}, 
      author={Takuya Akiba and Shotaro Sano and Toshihiko Yanase and Takeru Ohta and Masanori Koyama},
journal={arXiv preprint arXiv:1907.10902},
      year={2019},
}

@article{daulton2020differentiable,
  title={Differentiable expected hypervolume improvement for parallel multi-objective Bayesian optimization},
  author={Daulton, Samuel and Balandat, Maximilian and Bakshy, Eytan},
  journal={Proc. Adv. Neural Inf. Process.
Syst},
  volume={33},
  pages={9851--9864},
  year={2020}
}

@article{dosovitskiy2020image,
  title={An image is worth 16x16 words: Transformers for image recognition at scale},
  author={Dosovitskiy, Alexey and Beyer, Lucas and Kolesnikov, Alexander and Weissenborn, Dirk and Zhai, Xiaohua and Unterthiner, Thomas and Dehghani, Mostafa and Minderer, Matthias and Heigold, Georg and Gelly, Sylvain and others},
  journal={arXiv preprint arXiv:2010.11929},
  year={2020}
}

@INPROCEEDINGS{5206848,
  author={Deng, Jia and Dong, Wei and Socher, Richard and Li, Li-Jia and Kai Li and Li Fei-Fei},
  booktitle={Proc. IEEE Conf. Comput. Vis. Pattern Recognit.}, 
  title={ImageNet: A large-scale hierarchical image database}, 
  year={2009},
  volume={},
  number={},
  pages={248-255},
  doi={10.1109/CVPR.2009.5206848}}

@article{erak2025adaptive,
  title={Adaptive token merging for efficient transformer semantic communication at the edge},
  author={Erak, Omar and Alhussein, Omar and Abou-Zeid, Hatem and Bennis, Mehdi and Muhaidat, Sami},
  journal={arXiv preprint arXiv:2509.09955},
  year={2025}
}





\end{document}